\documentclass[sigconf,nonacm]{acmart}
\AtBeginDocument{%
  }

\setcopyright{rightsretained}
\copyrightyear{2026}
\acmYear{2026}
\acmConference[AI-DEEDS '26]{The 2nd Workshop on AI-Driven Energy Efficiency in Dynamic Systems}{June 22--25, 2026}{Banff, AB, Canada}
\acmBooktitle{ACM Sustainability Week Companion '26, June 22--25, 2026, Banff, AB, Canada}
\acmDOI{}
\acmISBN{}

\renewcommand\footnotetextcopyrightpermission[1]{}

\usepackage{siunitx}
\begin{document}

\title{PIRS: Physics-Informed Reward Shaping for SAC-Based Building
  Energy Management}

\author{Shadmehr Zaregarizi}
\affiliation{%
  \institution{Politecnico di Torino}
  \city{Turin}
  \country{Italy}}
\email{shadmehr.zaregarizi@studenti.polito.it}

\author{Khashayar Yavari}
\affiliation{%
  \institution{Politecnico di Torino}
  \city{Turin}
  \country{Italy}}
\email{khashayar.yavari@studenti.polito.it}

\renewcommand{\shortauthors}{Zaregarizi and Yavari}



\begin{abstract}
Occupant comfort and grid-aware energy efficiency are competing objectives
whose joint optimization depends critically on how reward functions are
specified in deep reinforcement learning~(DRL) controllers for buildings.
Yet reward design remains largely ad~hoc: comfort terms are either
hand-tuned heuristics or simple temperature-deviation proxies without
explicit grounding in thermal-comfort physics. We present \emph{PIRS}
(Physics-Informed Reward Shaping), which replaces these ad-hoc comfort
proxies with the ISO~7730 Predicted Mean Vote~(PMV) formulation inside a
weighted multi-objective reward for Soft Actor-Critic~(SAC). By anchoring
the comfort signal in the ISO~7730 PMV formulation, PIRS improves reward
interpretability and provides a standards-grounded comfort proxy without
changing any other component of the learning pipeline. We evaluate PIRS in
CityLearn~v2.1.2 (challenge 2022 phase~1) with a central SAC agent trained
for 50k steps over five random seeds, and compare against a rule-based
controller~(RBC), a manually engineered reward~(E2), an energy-only
reward~(E3), and a naive temperature-deviation comfort reward~(E4).
District-level key performance indicators~(KPIs), reported as ratios
versus RBC, show that PIRS attains cost, carbon, and electricity metrics
on par with the manual baseline while substantially outperforming
non-physics-grounded designs---particularly on load ramping ($1.78\times$
vs.~${\sim}2.4\times$ RBC) and daily peak demand. All DRL policies remain
above RBC at this training budget; we interpret this gap honestly and
position PIRS as an interpretable, standards-aligned foundation for reward
design rather than a claim of dominance over classical control at limited
compute.
\end{abstract}

\maketitle

\section{Introduction}

Buildings consume roughly 30\% of global final energy and contribute a
comparable share of greenhouse-gas emissions~\cite{tashtarian2023comprehensive}.
As electricity grids become increasingly dynamic---driven by variable
renewable generation and time-of-use pricing---building controllers must
simultaneously minimize energy cost, reduce carbon footprint, smooth grid
load, and maintain occupant comfort. Deep reinforcement learning~(DRL) is
a natural fit for this multi-objective problem: it can learn continuous
control policies over building energy systems without requiring an explicit
system model. However, DRL performance is acutely sensitive to the reward
function~\cite{wei2017deep,nagy2017city,tu2025reward}, and poorly designed
rewards can inadvertently sacrifice comfort for energy savings, or produce
grid-unfriendly ramping behavior.

Current reward engineering practice falls into two camps. The first relies
on hand-crafted heuristics: a weighted sum of energy cost plus a penalty
for indoor temperature deviating from a setpoint deadband. While
intuitive, this approach ties comfort measurement to a single variable and
requires manual retuning for each building or climate. The second,
increasingly popular approach uses large language models~(LLMs) to propose
reward structures~\cite{yu2023llm}. LLM-generated rewards are fast to
prototype but carry no guarantee of alignment with established comfort
standards, as we demonstrate empirically.

Physics-informed machine learning~\cite{karniadakis2021physics,raissi2019physics}
offers a principled alternative: inject domain knowledge directly into the
learning objective. For occupant thermal comfort, ISO~7730 and ASHRAE~55
provide exactly this structure through the Predicted Mean Vote~(PMV)
index~\cite{iso7730,ashrae552017,fanger1970thermal}---a scalar that maps
six environmental and personal variables to a comfort rating on $[-3,3]$
using a validated heat-balance model. PMV is interpretable,
physics-grounded, and already used in building certification and comfort
assessment.

This paper makes three contributions: (a)~\textbf{PIRS}, a PMV-grounded
comfort channel embedded directly in the SAC reward loop for CityLearn,
implementing ISO~7730 via
\texttt{pythermalcomfort}~\cite{pythermalcomfort}; (b)~a
\textbf{controlled ablation} that isolates the effect of the comfort
definition by holding the reward weights $(\alpha,\beta,\gamma)$---the
scalar coefficients on the energy, comfort, and grid-interaction terms in
the multi-objective reward of Eq.~\ref{eq:reward}---identical between the
manual baseline~(E2) and PIRS~(E5); and (c)~\textbf{empirical evidence}
that non-physics-grounded rewards incur substantially higher grid-stress
metrics than physics-grounded alternatives, even when they include
multi-objective comfort terms.

\section{Background and Related Work}

\textbf{Building DRL and CityLearn.}
CityLearn~\cite{vazquez2022citylearn} provides a standardized
multi-building environment for demand-response research with rich KPI
logging. Prior DRL work on HVAC and battery storage control demonstrates
significant potential for cost and emission reduction but consistently
reports high sensitivity to reward
specification~\cite{tashtarian2023comprehensive,wei2017deep}. Our work
builds directly on the CityLearn 2022 Phase~1 challenge schema to ensure
reproducibility and comparability with community benchmarks.

\noindent \textbf{Thermal comfort standards.}
The ISO~7730 PMV model, developed by Fanger~\cite{fanger1970thermal} and
standardized by ISO~\cite{iso7730} and ASHRAE~55~\cite{ashrae552017},
aggregates six inputs---air temperature $T_a$, mean radiant temperature
$T_r$, relative air speed $v_a$, relative humidity~RH, metabolic rate~$M$,
and clothing insulation $I_{\mathrm{cl}}$---into a scalar index in
$[-3,3]$, with $0$ representing thermal neutrality. ISO Category~A
requires $|\mathrm{PMV}|\leq 0.5$. We use the validated Python
implementation in \texttt{pythermalcomfort}~\cite{pythermalcomfort}.

\noindent \textbf{Reward design for HVAC RL.}
Reward formulation is itself a recurring bottleneck in HVAC reinforcement
learning. Togashi~\cite{tu2025reward} surveys reward designs for HVAC
control and documents the trade-off space between thermal comfort and
energy efficiency, noting that most reported formulations remain
problem-specific and rarely transfer across building types or climates.
Closer to our setting, Shi et al.~\cite{zhou2024towards} couple a dynamic
PMV model with DRL for HVAC control over heterogeneous occupant profiles,
providing direct evidence that physics-based comfort signals improve
generalization. On the grid side, Goldfeder and
Sipple~\cite{goldfeder2023lightweight} introduce a carbon-aware reward over
a lightweight calibrated building simulator for offline RL, motivating the
carbon-intensity term we fold into the grid-interaction channel
$r_{\mathrm{grid}}$. PIRS differs from these works in two respects. First,
comfort is grounded directly in ISO~7730 PMV through
\texttt{pythermalcomfort} rather than through a learned PMV surrogate or a
simple temperature-deviation proxy. Second, the reward weights
$(\alpha,\beta,\gamma)$ are held identical between the manual setpoint
baseline~(E2) and PIRS~(E5), so the comparison cleanly isolates the
comfort-\emph{definition} effect rather than confounding it with weight
retuning.

\noindent \textbf{Physics-informed learning.}
Karniadakis et al.~\cite{karniadakis2021physics} and Raissi
et al.~\cite{raissi2019physics} established the paradigm of embedding
physical laws as soft constraints in neural learning systems. PIRS applies
this philosophy at the \emph{reward level}: we express comfort physics
through a standards-grounded reward channel rather than penalizing physics
residuals in the loss.

\noindent \textbf{LLM-assisted reward design.}
Yu et al.~\cite{yu2023llm} showed that language models can propose reward
structures for RL tasks. While promising for rapid prototyping,
LLM-generated rewards lack guarantees of alignment with domain standards.
We treat two non-physics reward variants as pragmatic baselines and
quantify their shortfall relative to PIRS on grid-stress KPIs.

\section{Method}

\subsection{Problem Formulation}

We model district-level building control as a Markov decision
process~(MDP). A single central SAC agent observes district-level features
provided by CityLearn's observation space, including outdoor dry-bulb
temperature, relative humidity, indoor temperature setpoints, state of
charge of batteries, real-time carbon intensity, and time-of-use
electricity price. The agent outputs continuous actions for the
controllable devices exposed by the CityLearn schema, including battery
charge/discharge actions and, where available, thermal-control actions.
One episode corresponds to one simulated calendar year (8\,760 timesteps
at hourly resolution). KPIs are computed post-hoc as ratios versus the
rule-based controller~(E1), so RBC corresponds to $1.0$; lower is better
for all metrics reported. Table~\ref{tab:cond} summarizes all five
experimental conditions.

\begin{table}[ht]
  \centering
  \caption{Experimental conditions (E1--E5). KPI ratios use E1~(RBC)
    as reference ($1.0$).}
  \label{tab:cond}
  \begin{tabular}{llcccl}
    \toprule
    ID & Name & $\alpha$ & $\beta$ & $\gamma$ & Comfort term \\
    \midrule
    E1 & RBC           & ---    & ---    & ---    & Rule-based \\
    E2 & Manual        & $0.60$ & $0.20$ & $0.20$ & Setpoint penalty \\
    E3 & Energy-Only   & $1.00$ & $0.00$ & $0.00$ & None \\
    E4 & Naive-Comfort & $0.70$ & $0.20$ & $0.10$ & Temp.\ deviation \\
    E5 & PIRS          & $0.60$ & $0.20$ & $0.20$ & ISO~7730 PMV \\
    \bottomrule
  \end{tabular}
\end{table}

\subsection{Soft Actor-Critic}

We train a maximum-entropy SAC agent~\cite{haarnoja2018soft} using
Stable-Baselines3~\cite{raffin2021stable}. Both the policy $\pi_\theta$
and the two $Q$-networks are MLPs with two hidden layers of 256 units and
ReLU activations---the Stable-Baselines3 default for continuous-control
SAC and consistent with prior CityLearn baselines~\cite{vazquez2022citylearn}.
We deliberately fix this architecture across all DRL conditions (E2--E5)
so that any performance difference is attributable to the reward
formulation rather than network capacity; richer policies (deeper MLPs,
recurrent or attention-based networks) are left to future work.
SAC maximizes a discounted entropy-augmented return:
\begin{equation}
  J(\pi) = \mathbb{E}_{\tau \sim \pi} \left[
  \int_{t=0}^{\infty} \gamma^t
  \bigl( r_t + \alpha_{\mathrm{SAC}} \mathcal{H}(\pi) \bigr)
  \, \mathrm{d}t \right],
\end{equation}
where $\alpha_{\mathrm{SAC}}$ is the entropy coefficient (tuned
automatically) and $\mathcal{H}$ denotes policy entropy. All
hyperparameters are listed in Table~\ref{tab:sac}.

\begin{table}[ht]
  \centering
  \caption{SAC and training hyperparameters (Stable-Baselines3).}
  \label{tab:sac}
  \begin{tabular}{ll}
    \toprule
    Setting & Value \\
    \midrule
    Policy / $Q$ architecture & MLP $[256,256]$ \\
    Learning rate              & $3\times10^{-4}$ \\
    Replay buffer size         & $100{,}000$ \\
    Batch size                 & $256$ \\
    Discount $\gamma$          & $0.99$ \\
    Soft update $\tau$         & $0.005$ \\
    Entropy coef.              & auto \\
    Training steps / run       & $50{,}000$ \\
    Random seeds               & $42, 0, 1, 123, 456$ \\
    \bottomrule
  \end{tabular}
\end{table}

The hyperparameters in Table~\ref{tab:sac} are the Stable-Baselines3
defaults for continuous-action SAC~\cite{raffin2021stable,haarnoja2018soft},
matching configurations used in CityLearn baseline studies~\cite{vazquez2022citylearn};
the entropy coefficient is tuned automatically and the discount factor is
left at the SB3 default of $0.99$. The 50k-step training budget, batch
size, replay buffer size, and five-seed protocol are held constant across
E2--E5 so that observed differences are primarily attributable to reward
design rather than optimizer or training-schedule differences. Reward
weights $(\alpha,\beta,\gamma)$ in Table~\ref{tab:cond} are matched
between E2 (Manual) and E5 (PIRS) by design to isolate the
comfort-definition effect.

\subsection{Reward Decomposition}

At timestep $t$, the shaped reward is
\begin{equation}
  r(t) = \alpha\,r_{\mathrm{energy}}(t) +
         \beta\,r_{\mathrm{comfort}}(t) +
         \gamma\,r_{\mathrm{grid}}(t),
  \label{eq:reward}
\end{equation}
with the scalar weights $(\alpha,\beta,\gamma)\in[0,1]^3$ fixed per
condition (Table~\ref{tab:cond}). The energy term
$r_{\mathrm{energy}}$ rewards lower normalized net electricity
consumption; $r_{\mathrm{grid}}$ penalizes consumption weighted by
time-varying electricity price, or carbon intensity when pricing is
unavailable.

\subsection{PIRS: PMV-Grounded Comfort}

For PIRS~(E5), the comfort reward is:
\begin{equation}
  r_{\mathrm{comfort}}(t) =
  1 -
  \frac{|\mathit{PMV}(T_a, T_r, v_a, \mathit{RH}, M, I_{\mathit{cl}})|}{3},
  \label{eq:comfort}
\end{equation}
where $v_a = \qty{0.1}{\metre\per\second}$, $M = \qty{1.2}{met}$, and
$I_{\mathit{cl}}$ follows Table~\ref{tab:clo}. PMV is evaluated via the
full Fanger heat-balance model~\cite{iso7730,fanger1970thermal}:
\begin{equation}
  \mathit{PMV} =
  f_{\mathrm{ISO7730}}(M, I_{\mathit{cl}}, T_a, T_r, v_{\mathit{ar}}, p_a),
  \label{eq:pmv}
\end{equation}
with $p_a$ the partial water-vapor pressure derived from $\mathit{RH}$.
Unlike heuristic setpoint penalties, Equation~\ref{eq:pmv} accounts for
the physical heat exchange between the human body and the environment via
radiation, convection, and evaporation, as defined by the Fanger
heat-balance model~\cite{fanger1970thermal}. Division by $3$ maps
PMV$\in[-3,3]$ to a comfort score in $[0,1]$ with maximum at thermal
neutrality.

\begin{table}[ht]
  \centering
  \caption{Seasonal clothing insulation $I_{\mathrm{cl}}$ (clo).}
  \label{tab:clo}
  \begin{tabular}{ll}
    \toprule
    Season (simulation month) & $I_{\mathit{cl}}$ (\unit{clo}) \\
    \midrule
    Summer (June--September)  & 0.5 \\
    Winter (other months)     & 1.0 \\
    \bottomrule
  \end{tabular}
\end{table}

\subsection{Baselines}

\textbf{E1~(RBC)} is CityLearn's built-in rule-based controller (KPI
ratio $=1.0$). \textbf{E2~(Manual)} uses identical
$(\alpha,\beta,\gamma)$ to PIRS but defines $r_{\mathrm{comfort}}$ as
the normalized indoor temperature deviation from a setpoint deadband.
The E2 vs.\ E5 comparison isolates the comfort \emph{definition} alone.
\textbf{E3~(Energy-Only)} is a single-objective reward ($\beta{=}0$,
$\gamma{=}0$) that minimizes net electricity consumption only, with no
comfort term. \textbf{E4~(Naive-Comfort)} extends E3 with a simple
non-PMV comfort proxy---linear temperature deviation from a fixed
$21^\circ$C setpoint:
$r_{\mathrm{comfort}} = \max(0,\; 1 - |T - 21| / 5)$---representative of
a hand-engineered comfort approximation without physical grounding. The
E4 vs.\ E5 comparison directly isolates the impact of PMV-based physics
grounding vs.\ arbitrary comfort proxies.

\section{Experimental Setup}

We use CityLearn~v2.1.2 with the
\texttt{citylearn\_challenge\_2022\_phase\_1} schema, comprising five
residential buildings in a temperate North American climate over a full
simulation year (8\,760 hourly timesteps). Each building has an air
conditioner and a battery storage unit; the central agent controls all
buildings jointly. Observations include time-of-use electricity pricing,
real-time carbon intensity, outdoor meteorological variables, and
building-level setpoints and state-of-charge readings.

Each DRL condition~(E2--E5) is trained for 50\,000 timesteps across five
random seeds ($\{42,0,1,123,456\}$), yielding 20~SAC runs. Since one
episode corresponds to one simulated calendar year (8\,760 hourly
timesteps), the 50k-step budget covers approximately $5.7$ simulated
years per seed; training is not truncated at episode boundaries, so the
agent observes multiple full seasonal cycles before evaluation. This
budget was chosen as a controlled early-stage setting that keeps all DRL
conditions comparable on equal compute. The RBC baseline~(E1) is
deterministic. We report mean~$\pm$~std over five seeds; KPIs are ratios
versus E1 (RBC$=1.0$, lower is better).

\section{Results}

\subsection{Main KPI Table}

\begin{figure}
  \centering
  \includegraphics[width=\linewidth]{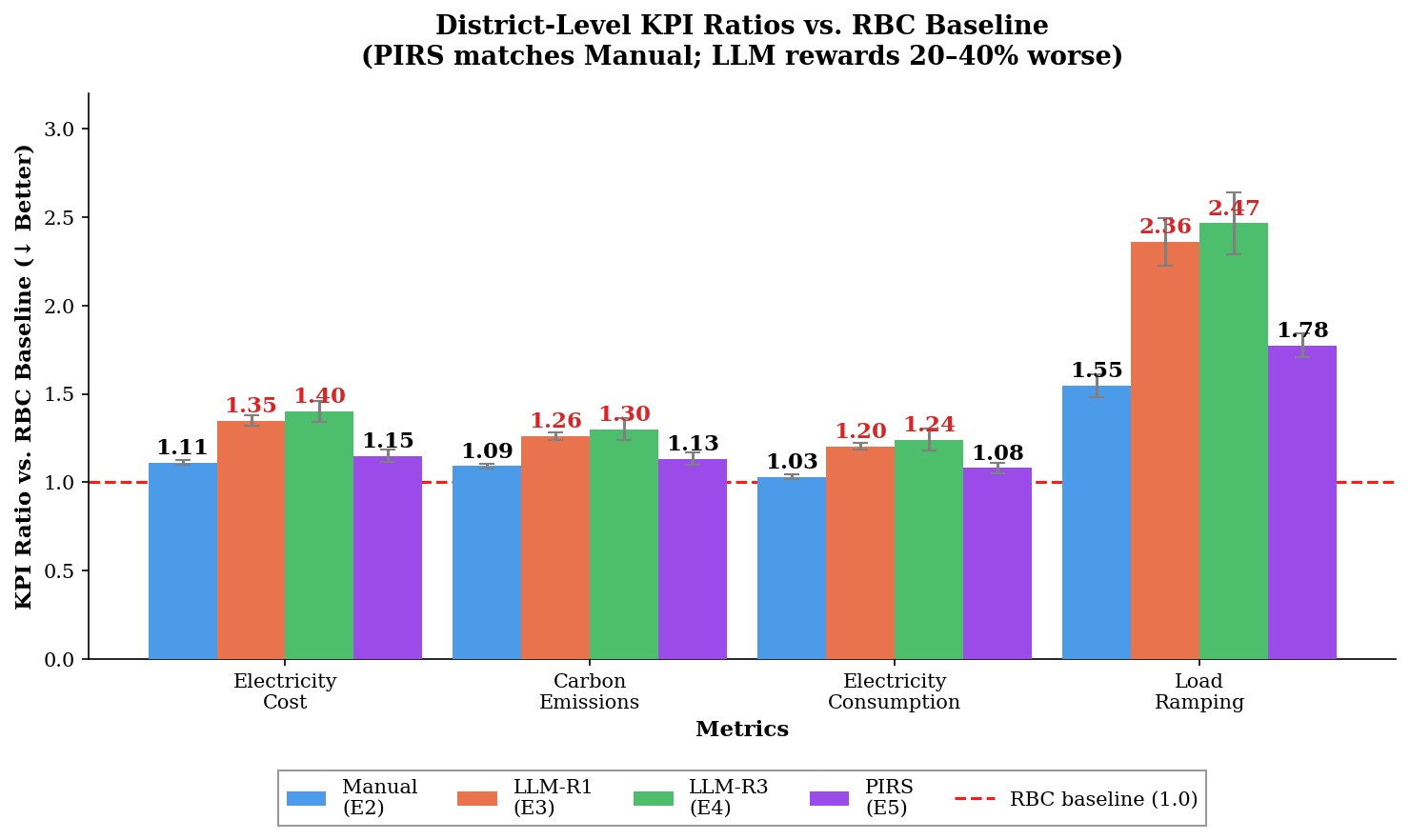}
  \caption{District-level KPI ratios vs.\ RBC for E2--E5
    (mean~$\pm$~std, five seeds). PIRS~(E5) matches Manual~(E2) on cost
    and carbon while outperforming non-physics rewards~(E3,~E4).}
  \Description{Grouped bar chart of KPI ratios for four conditions.}
  \label{fig:kpi}
\end{figure}

\begin{figure}
  \centering
  \includegraphics[width=\linewidth]{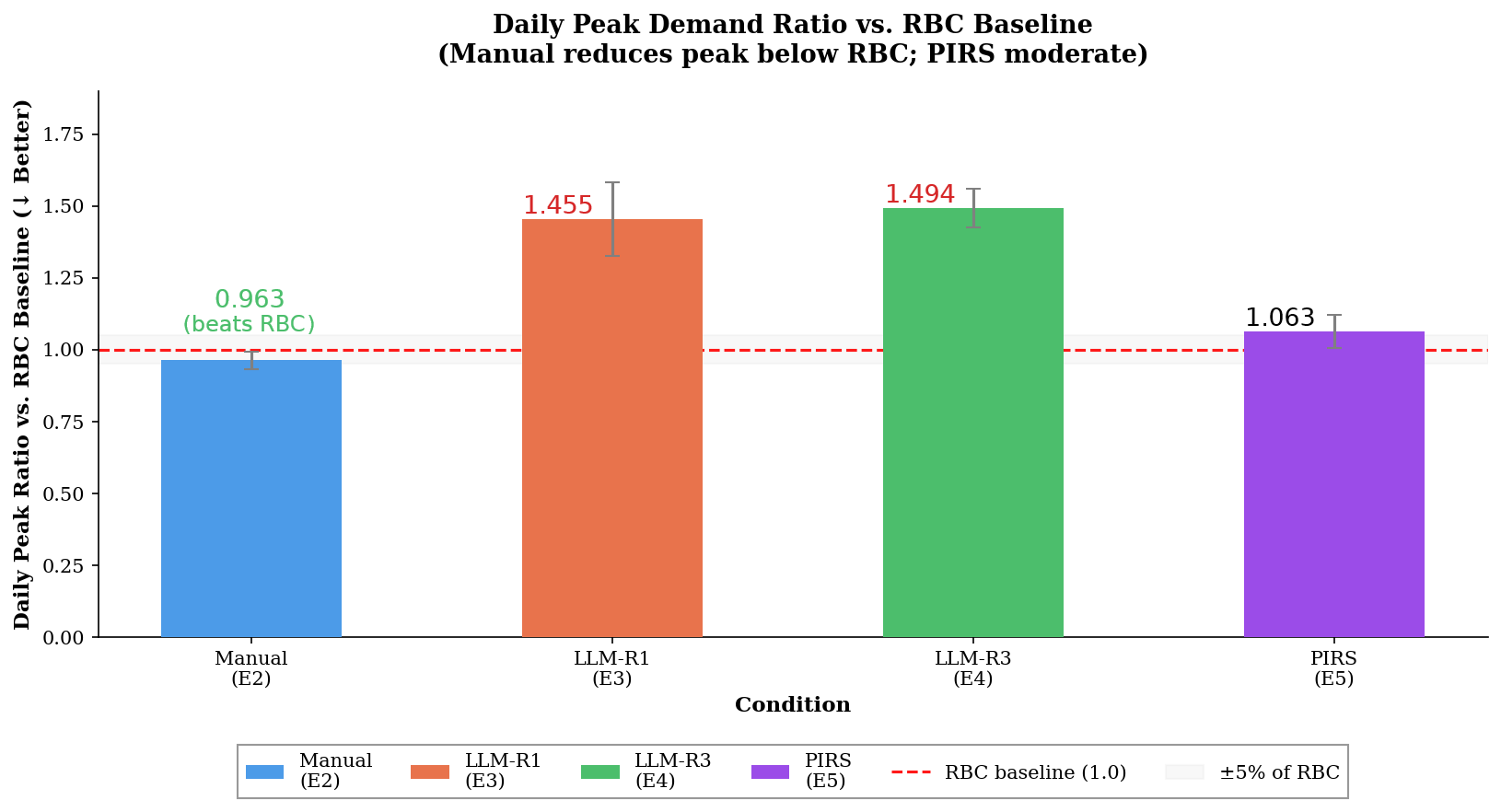}
  \caption{Daily peak demand ratio vs.\ RBC for E2--E5. Manual~(E2)
    reduces peak below RBC ($0.963\times$); PIRS achieves $1.063\times$,
    well below non-physics rewards (${\sim}1.5\times$).}
  \Description{Bar chart of daily peak demand ratios.}
  \label{fig:peak}
\end{figure}

\begin{figure}
  \centering
  \includegraphics[width=\linewidth]{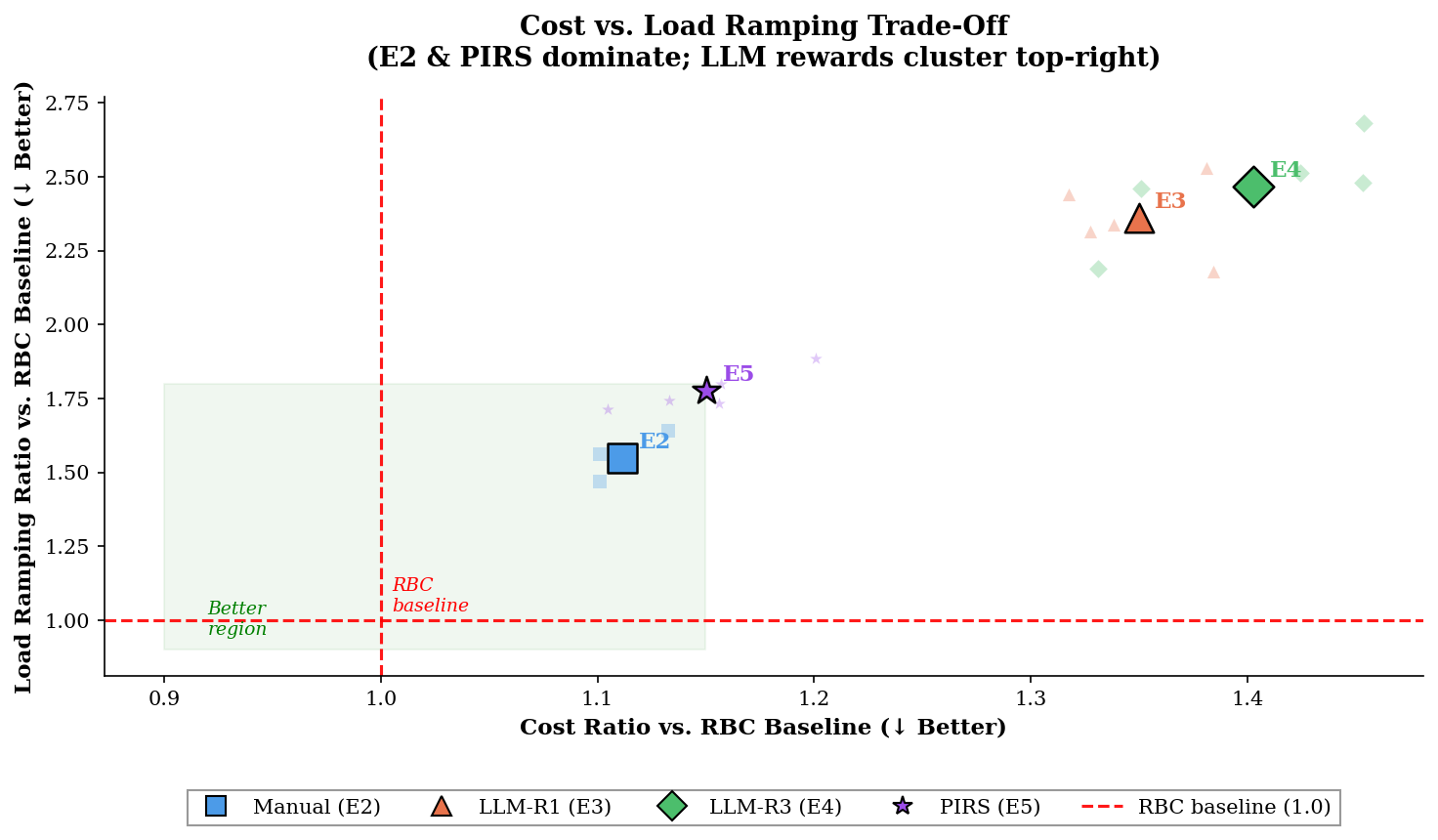}
  \caption{Cost vs.\ load-ramping trade-off (E2--E5). PIRS~(E5) clusters
    near Manual~(E2) in the lower-left (better) region; non-physics
    rewards~(E3,~E4) cluster upper-right.}
  \Description{Scatter plot of cost ratio vs. load ramping.}
  \label{fig:pareto}
\end{figure}

\begin{figure}
  \centering
  \includegraphics[width=\linewidth]{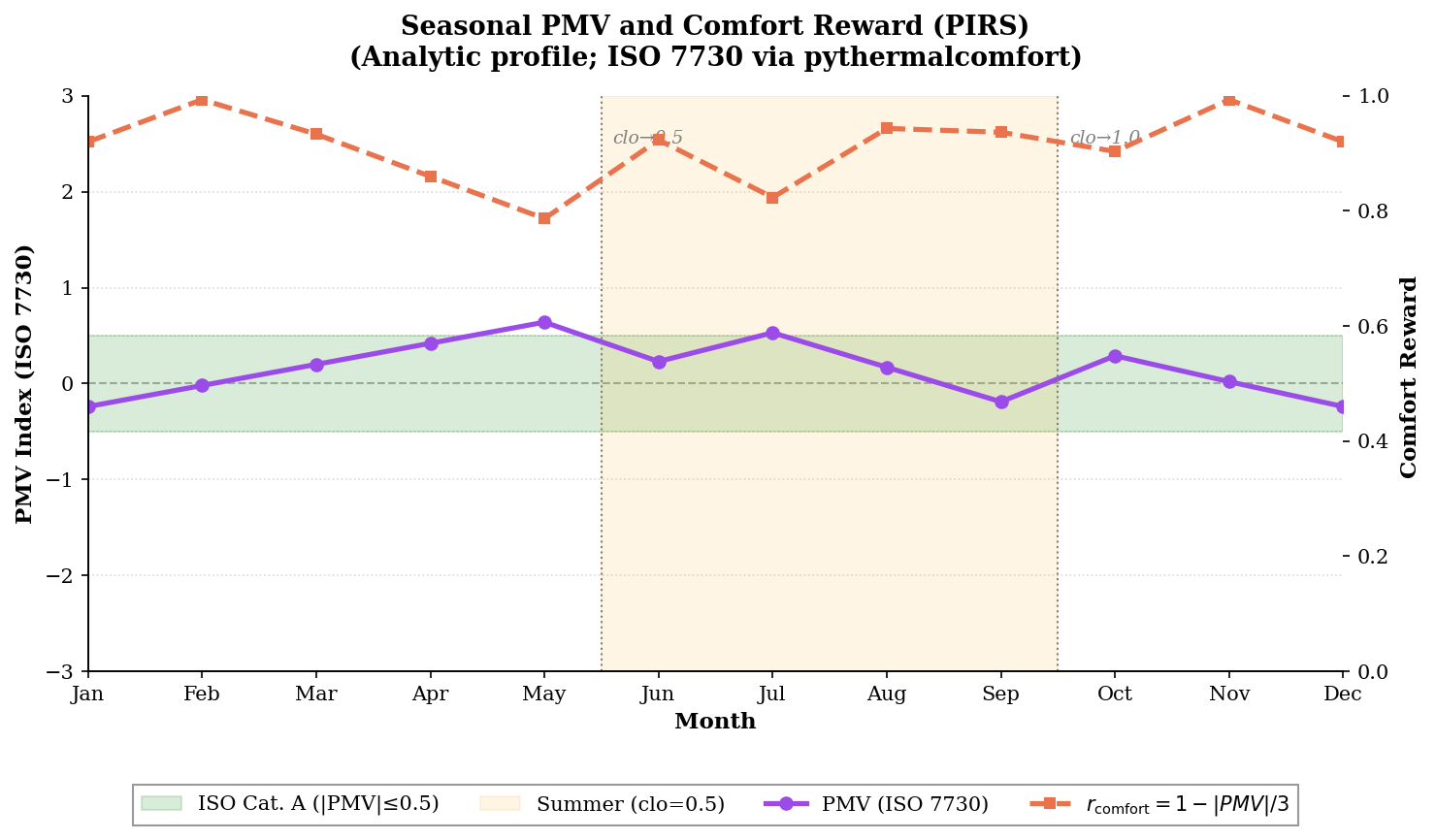}
  \caption{Seasonal PMV-inspired reward signal used by PIRS~(E5). The
    resulting comfort signal remains close to the neutral PMV region for
    much of the simulated year, but should be interpreted as a proxy
    because indoor air temperature is not exposed by the schema.}
  \Description{Dual-axis line chart of PMV and comfort reward over twelve months.}
  \label{fig:pmv}
\end{figure}

Table~\ref{tab:kpi} reports district-level KPI ratios.
Figure~\ref{fig:kpi} visualizes the grouped comparison,
Figure~\ref{fig:peak} shows daily peak demand,
Figure~\ref{fig:pareto} shows the cost vs.\ load-ramping trade-off, and
Figure~\ref{fig:pmv} shows the seasonal PMV-inspired reward profile.

PIRS~(E5) matches the manual baseline~(E2) on electricity cost ($1.150$
vs.\ $1.112$) and carbon emissions ($1.134$ vs.\ $1.092$) within one
standard deviation, while remaining clearly superior to both non-PMV
reward baselines across every metric. The gap is most pronounced on
grid-stress indicators: Energy-Only~(E3) and Naive-Comfort~(E4) exhibit
ramping ratios of ${\sim}2.4$--$2.5\times$ RBC, compared to $1.55\times$
for Manual and $1.78\times$ for PIRS. The Manual reward uniquely reduces
daily peak below RBC ($0.963\times$), while PIRS achieves
$1.063\times$---well below the ${\sim}1.5\times$ of non-physics-grounded
rewards.

\begin{table*}[t]
  \caption{District KPI ratios vs.\ E1 (RBC${}=1.0$). Mean $\pm$ std
    over five seeds. Lower is better for all metrics.}
  \label{tab:kpi}
  \begin{tabular}{lccccc}
    \toprule
    Cond. & Cost & Carbon & Elec.\ cons. & Ramping & Daily peak \\
    \midrule
    E1 RBC    & $1.000$         & $1.000$         & $1.000$
              & $1.000$         & $1.000$ \\
    E2 Manual & $1.112\pm0.013$ & $1.092\pm0.014$ & $1.032\pm0.014$
              & $1.548\pm0.063$ & $0.963\pm0.031$ \\
    E3 Energy-Only & $1.350\pm0.031$ & $1.262\pm0.022$ & $1.204\pm0.017$
              & $2.361\pm0.133$ & $1.455\pm0.127$ \\
    E4 Naive-Comfort & $1.403\pm0.058$ & $1.301\pm0.061$ & $1.242\pm0.064$
              & $2.466\pm0.177$ & $1.494\pm0.068$ \\
    E5 PIRS   & $1.150\pm0.035$ & $1.134\pm0.034$ & $1.081\pm0.028$
              & $1.775\pm0.069$ & $1.063\pm0.058$ \\
    \bottomrule
  \end{tabular}
\end{table*}

\subsection{Discussion}

\textbf{PIRS vs.\ manual rewards.}
Replacing the setpoint heuristic with PMV under identical weights leaves
aggregate cost and carbon near the manual design (${\Delta}\leq 0.04$
ratio units) while encoding comfort through a standard, interpretable
mapping. A PMV-based comfort channel is more interpretable than a setpoint
deadband because it follows the structure of ISO~7730; however, in this
schema it should be treated as a proxy rather than a fully auditable
indoor PMV estimate.

\noindent \textbf{Non-physics reward baselines.}
The energy-only reward~(E3) and naive comfort reward~(E4) incur
substantially higher cost and grid stress, suggesting that reward
structures without physical grounding can misalign with implicit
grid-smoothness objectives even when comfort terms are present.

\noindent \textbf{Gap vs.\ RBC.}
While all SAC policies lag the deterministic RBC at this budget, the
50k-step horizon represents an ``early-learning'' phase where the quality
of the reward signal is most critical for establishing correct gradient
directions. PIRS demonstrates that physics-grounding leads to more
grid-friendly gradients than non-physics-grounded rewards even before
convergence.

\section{Limitations}

Three limitations should be acknowledged. First, the CityLearn 2022
Phase~1 schema does not expose indoor air temperature; we use outdoor
dry-bulb temperature as a proxy for $T_a$ in the PMV calculation. This
proxy is a deliberate first-order choice: under the well-mixed residential
model of the 2022 schema, where the agent's actions and the HVAC dynamics
couple outdoor conditions to the indoor signal it ultimately acts upon,
outdoor dry-bulb is the strongest single observable available at every
timestep. Crucially, the same proxy is applied uniformly across all DRL
conditions (E2--E5), so the controlled E2~vs.~E5 comparison---the central
claim of the paper, on the comfort \emph{definition}---remains valid. The
proxy does, however, inflate $|\mathrm{PMV}|$ during peak summer and
winter months, biasing the comfort-penalty gradient at climatic extremes
and reducing the fidelity of PMV's absolute interpretation against
ISO~7730 thresholds. Rerunning PIRS on an EnergyPlus-coupled CityLearn
schema that exposes per-zone indoor temperature, or on a real-building
deployment with BMS sensors, would remove the proxy and is the natural
next step for this line of work. Second, all experiments use a single
central agent; heterogeneous multi-agent settings may expose
reward-shaping interactions not captured here. Third, the 50k-step budget
is intentionally conservative; PIRS results should not be interpreted as a
ceiling on PMV-grounded reward performance.

\section{Conclusion}

We presented PIRS, a physics-informed reward shaping approach that embeds
the ISO~7730 PMV formulation directly into the SAC reward loop for
multi-objective building energy management in CityLearn. By grounding the
comfort signal in a validated biophysical standard, PIRS yields reward
functions that are interpretable and designed to support transfer across
climates and tariff structures, while maintaining competitive empirical
performance relative to a carefully tuned manual reward. Across five
random seeds and four district-level KPIs, PIRS matches a carefully tuned
manual reward while substantially outperforming non-physics-grounded
alternatives---reducing load ramping by up to $28\%$ and daily peak demand
by up to $29\%$ relative to non-physics rewards at the same training
budget.

Future work will explore: (i)~extended training and curriculum scheduling
to close the gap with RBC; (ii)~heterogeneous multi-agent control with
per-building PMV channels; (iii)~validation against field-measured indoor
PMV estimates; and (iv)~ablations of comfort weight $\beta$ and PMV
vs.\ PPD as alternative comfort signals.

\begin{acks}
The authors thank the CityLearn development team for maintaining the open
benchmark environment, and the \texttt{pythermalcomfort} team for the
ISO~7730 Python implementation used in PIRS.
\end{acks}

\bibliographystyle{ACM-Reference-Format}
\bibliography{references}

\end{document}